\documentclass[runningheads]{llncs}
\usepackage{graphicx}
\usepackage{graphicx}
\usepackage{enumitem}

\usepackage{times}
\usepackage{latexsym}

\usepackage{url}

\usepackage[dvipsnames]{xcolor}
\usepackage{multirow}
\usepackage{array}
\usepackage{boldline}

\usepackage{amsmath}

\begin{document}
%
\title{Recent Advances in End-to-End Spoken Language Understanding}

\titlerunning{Recent Advances in End-to-End Spoken Language Understanding}

\author{Natalia Tomashenko\inst{1} \and
Antoine Caubri\`ere\inst{2} \and
Yannick Est\`eve\inst{1} \and
 Antoine Laurent\inst{2} \and
 Emmanuel Morin\inst{3}}
\vspace{-20pt}

\authorrunning{N. Tomashenko et al.}
%
\institute{LIA, University of Avignon, France \\
\email{\{natalia.tomashenko, yannick.esteve\}@univ-avignon.fr} \\
\and
LIUM, University of Le Mans, France\\
\email{\{antoine.caubriere,antoine.laurent\}@univ-lemans.fr}\and
LS2N,  University of Nantes\\
\email{emmanuel.morin@univ-nantes.fr}}
\maketitle              
\vspace{-10pt}
\begin{abstract}
  This work investigates  spoken language understanding (SLU) systems in the scenario when the semantic information is extracted directly from the  speech signal by means of a single end-to-end neural network model.
  Two SLU tasks are considered:  named entity recognition  (NER) and semantic slot filling (SF). For these tasks, in order  to improve the model performance, we explore  various techniques  including  speaker adaptation,  a modification of the connectionist temporal classification (CTC) training criterion,  and sequential pretraining. 
\end{abstract}
\section{Introduction}\label{sec:intro}

Spoken language understanding (SLU) is a key component of  conversational artificial intelligence (AI) applications.
Traditional SLU systems consist of at least two parts. The first one is an automatic speech recognition (ASR) system that transcribes acoustic speech signal into word sequences. The second part is a natural language understanding (NLU) system which predicts, given the output of the ASR system, named entities, semantic or domain  tags, and other language characteristics depending on the considered task.  In  classical approaches, these two systems are often built and optimized independently.

Recent progress in deep learning has impacted  many research and industrial domains and  boosted the development of conversational AI technology.
Most of the state-of-the art SLU and conversational AI systems employ neural network models.  
Nowadays there is a high interest of the research community  in end-to-end systems for various speech and language technologies.
A few recent papers~\cite{qian2017exploring,haghani2018audio,serdyuk2018towards,ghannay2018end,chen2018spoken,lugosch2019speech} present 
ASR-free end-to-end approaches for SLU tasks and show promising results.
These methods aim to learn SLU models from acoustic signal without  intermediate text representation.
Paper~\cite{chen2018spoken} proposed an audio-to-intent architecture for semantic classification in dialog systems.
An encoder-decoder framework~\cite{sutskever2014sequence} is used in~\cite{serdyuk2018towards} for domain and intent classification, and in~\cite{haghani2018audio} for domain, intent, and argument recognition. A different approach based on the model trained with 
the connectionist temporal classification (CTC) criterion~\cite{graves2006connectionist} was proposed in~\cite{ghannay2018end} for named entity recognition (NER) and slot filling. 
End-to-end methods are motivated by the following factors:
possibility of better information transfer from the speech signal due to the joint optimization  on the final objective criterion,
and simplification of the overall system and elimination of some of its components.
However, deep neural networks and especially end-to-end models often require more training data to be efficient.
For SLU, this implies the demand of big semantically annotated corpora. 
In this work, we explore different ways to improve the performance  of end-to-end SLU systems.

\section{SLU tasks}\label{sec:slu_tasks}

In SLU for human-machine conversational  systems, an important task is
to automatically extract semantic concepts or to fill in a set of \textit{slots}  in order to
achieve a goal in a human-machine dialogue. 
In this paper, we consider two SLU tasks: named entity recognition (NER) and semantic slot filling (SF).
In the NER task, the purpose is to recognize information units such as names, including person, organization and location names, dates, events and others. 
In the SF task, the extraction of
 wider semantic information is targeted. 
These last years, NER and SF where addressed as word labelling problems, through the use of  the classical BIO \textit{(begin/inside/outside)} notation. 
For instance, "\textit{I would like to book three double rooms in Paris for tomorrow}" will be represented for the NER and SF task as the following BIO labelled sentences:
\begin{itemize}
\vspace{-3pt}
\footnotesize
\item NER: "\textit{I\textcolor{blue}{::$\emptyset$} would\textcolor{blue}{::$\emptyset$} like\textcolor{blue}{::$\emptyset$} to\textcolor{blue}{::$\emptyset$} book\textcolor{blue}{::$\emptyset$} three\textcolor{blue}{::B-amount} double::\textcolor{blue}{$\emptyset$} rooms\textcolor{blue}{::$\emptyset$} in::\textcolor{blue}{$\emptyset$} Paris\textcolor{blue}{::B-location/city} for\textcolor{blue}{::$\emptyset$} tomorrow\textcolor{blue}{::B-time/date}}". 
\item SF: "\textit{I\textcolor{blue}{::B-command} would\textcolor{blue}{::I-command} like\textcolor{blue}{::I-command} to\textcolor{blue}{::I-command} book\textcolor{blue}{::I-command} three\textcolor{blue}{::B-room/number} double\textcolor{blue}{::B-room/type} rooms\textcolor{blue}{::I-room/type} in\textcolor{blue}{::$\emptyset$} Paris\textcolor{blue}{::B-location/city} for\textcolor{blue}{::$\emptyset$} tomorrow\textcolor{blue}{::B-time/date}}".
\end{itemize}

In this paper, similarly to~\cite{ghannay2018end}, the BIO representation is abandoned in profit to a chunking approach. 
For instance for NER, the same sentence will be presented as: 
\begin{itemize}
\footnotesize
    \item NER: "\textit{I would like to book} $<$ \textcolor{blue}{\textit{amount}} \textit{three} $>$ \textit{double rooms in} $<$ \textcolor{blue}{\textit{location/city}} \textit{Paris} $>$ \textit{for} $<$ \textcolor{blue}{\textit{time/date}} \textit{tomorrow} $>$".
\end{itemize}


In this study, we train an end-to-end neural model to reproduce such textual representation from speech.
Since our neural model emits characters, we use specific characters corresponding to each opening tag (one by named entity category or one by semantic concept), while the same symbol is used to represent the closing tag.


\section{Model training}\label{sec:model-training}

End-to-end training of SLU models is realized through the recurrent neural network (RNN) architecture and CTC loss function~\cite{graves2006connectionist} as shown in Figure~\ref{fig:ds}.
A spectrogram of power normalized audio clips  calculated on 20ms windows is used as the input features for the system.
As shown in Figure~\ref{fig:ds}, it is followed by two 2D-invariant (in the time and-frequency domain) convolutional layers, and then by five BLSTM layers with sequence-wise batch normalization.
A fully connected layer is applied after BLSTM layers, and  the output layer of the neural network  is a softmax layer.  
The model is trained using the CTC loss function.
The neural architecture is  similar to the Deep Speech 2~\cite{amodei2016deep} for ASR.

The outputs of the network depend on the task. 
For ASR, the outputs consist of  graphemes of a corresponding language, a \textit{space} symbol to denote word boundaries and a \textit{blank} symbol. For NER, in addition to ASR outputs, we add outputs corresponding to named entity types and a closing symbol for named entities.
In the same way, for SF, we use all ASR outputs and additional tags corresponding to semantic concepts and a closing symbol for semantic tags.

In order to improve model training, we investigate speaker adaptive training (SAT), pretraining and transfer learning. First, we formalize the $\star$-\textit{mode}, that proved its effectiveness in all our previous and current experiments.

\subsection{CTC loss function interpretation related to \textit{$\star$-mode} }\label{sec:starmode}

The CTC loss function~\cite{graves2006connectionist} is  relevant to train models for ASR without  Hidden Markov Models. The \textit{$\star$-mode} can be seen as a minor modification of the CTC loss function.

\subsubsection{CTC loss function definition}

By means of a many-to-one $\mathcal{B}$ mapping function, CTC transforms a sequence of the network outputs, emitted for each acoustic frame, to a sequence of final target labels by deleting repeated output labels and  inserting a \textit{blank} (\textit{no label}) symbol. 
The CTC loss function is defined as:
\begin{equation}
\mathcal{L}_{CTC} = - \sum_{(\mathbf{x}, \mathbf{l})\in Z} \ln P(\mathbf{l}|\mathbf{x}),
\end{equation}
where $\mathbf{x}$ is a sequence of acoustic observations, $\mathbf{l}$ is the target output label sequence,
and $Z$ the training dataset.
$P(\mathbf{l}|\mathbf{x})$ is defined as:
\begin{equation}
\label{eq:Plx}
P(\mathbf{l}|\mathbf{x})=\sum_{\pi \in \mathcal{B}^{-1}(\mathbf{l})} P(\pi|\mathbf{x}),
\end{equation}
where 
$\pi$ is a sequence of initial output labels emitted by the model for each input frame. To compute $P(\pi|\mathbf{x})$ we use the probability of the output label $\pi_t$ emitted by the neural model for frame $t$ to build this sequence. This probability is modeled by the value  $y_{\pi_t}^t$ given by the output node of the neural model related to the label $\pi_t$.
$P(\pi|\mathbf{x})$ is defined as
$
P(\pi|\mathbf{x})=\prod_t^T y_{\pi_t}^t,
$
where $T$ denotes the number of frames.

\subsubsection{CTC loss function and \textit{$\star$-mode} }

In the framework of the \textit{$\star$-mode}, we introduce a new symbol, "\textit{$\star$}", that represents the presence of a label (the opposite of the \textit{blank} symbol) that does not need to be disambiguated. 
We expect to build a model that is more discriminant on the important task-specific labels. 
For example, for the SF SLU task important labels are the ones corresponding to semantic concept  opening and closing tags, and characters involved in the word sequences that support the value  of these semantic concepts (\textit{i.e} characters occurring between an opening and a closing concept tag).
In the CTC loss function framework, the \textit{$\star$-mode} consists in applying another kind of mapping function before  $\mathcal{B}$. 
While $\mathcal{B}$ converts a sequence $\pi$ of initial output labels  into the final sequence $\mathbf{l}$ to be retrieved, we introduce the mapping function $\mathcal{S}$  that is applied to each final target output label.
Let $C$ be the set of elements $\mathbf{l}_i$ included in subsequences $\mathbf{l}_a^b \subset \mathbf{l}$ such as $\mathbf{l}_a$ is an opening concept tag and $\mathbf{l}_b$ the associated closing tag;  $i$, $a$ and $b$ are indexes that handle positions in sequence $\mathbf{l}$, and $a\leq i \leq b$. 
Let $V$ be the vocabulary of all the symbols present in sequences $\mathbf{l}$ in $Z$, and let consider the new symbol $\star \notin V$. Let define $V^\star =V \cup \left\{ \star \right\}$, and $L$ (resp. $L^\star$) the set of all the label sequences that can be generated from $V$ (resp. $V^\star$).

Considering $n$ as the number of elements in $\mathbf{l}$, $m$ an integer such as $m \leq n$, we define the mapping function $\mathcal{S}:L \rightarrow L^\star, \mathbf{l} \mapsto \mathbf{l^\prime}$  in two steps: 
\vspace{-2pt}
\begin{equation}
\begin{array}{llc}
1. & \forall \mathbf{l}_j \in \mathbf{l} &
\begin{cases}
    \mathbf{l}_j\notin C \Rightarrow \mathbf{l^\prime}_j=\star   \\
    \mathbf{l}_j\in C \Rightarrow \mathbf{l^\prime}_j=\mathbf{l}_j \\ 
\end{cases}\\
2. &  \forall \mathbf{l^\prime}_j \in \mathbf{l^\prime} & \mathbf{l^\prime}_{j-1} = \star \Rightarrow \mathbf{l^\prime}_j = \emptyset
\end{array}
\end{equation}

By applying $\mathcal{S}$ on the last example sentence used in Section~\ref{sec:slu_tasks} for NER, this sentence is transformed to:
\begin{itemize}
\vspace{-3pt}
\footnotesize
    \item sentence: "\textit{I would like to book} $<$ \textcolor{blue}{\textit{amount}} \textit{three} $>$ \textit{double rooms in} $<$ \textcolor{blue}{\textit{location/city}} \textit{Paris} $>$ \textit{for} $<$\textcolor{blue}{\textit{time/date}} \textit{tomorrow} $>$".
    \item $\mathcal{S}$(sentence): "\textit{\textcolor{purple}{{$\star$}}} $<$ \textcolor{blue}{\textit{amount}} \textit{three} $>$ \textit{\textcolor{purple}{$\star$}} $<$ \textcolor{blue}{\textit{location/city}} \textit{Paris} $>$ \textit{\textcolor{purple}{$\star$}} $<$ \textcolor{blue}{\textit{time/date}} \textit{tomorrow} $>$".
\end{itemize}

To introduce \textit{$\star$-mode} in the CTC loss function definition, we modify the formulation of $P(\mathbf{l}|\mathbf{x})$ in formula (\ref{eq:Plx}) by introducing the $\mathcal{S}$ mapping function applied to~$\mathbf{l}$:
\begin{equation}
\label{eq:PlxSTAR}
P(\mathbf{l}|\mathbf{x})=\sum_{\pi \in \mathcal{B}^{-1} \circ \mathcal{S} (\mathbf{l})} P(\pi|\mathbf{x}).
\end{equation}

\vspace{-10pt}
\subsection{Speaker adaptive training}\label{sec:speak_adapt}

Adaptation is an efficient way to reduce the
mismatches between the models and the data from a particular
speaker or channel. For many years, acoustic model adaptation has been a key component of any state-of-the-art ASR system.
%
For end-to-end approaches, speaker adaptation is less studied, and most of the first end-to-end ASR systems do not use any speaker adaptation and are built on 
spectrograms~\cite{amodei2016deep} or filterbank features~\cite{bahdanau2016end}.
However, some recent works~\cite{delcroix2018auxiliary,tomashenko2018evaluation} 
have demonstrated the effectiveness of speaker adaptation for end-to-end models.

\begin{figure}
\centering\includegraphics[width=72mm]{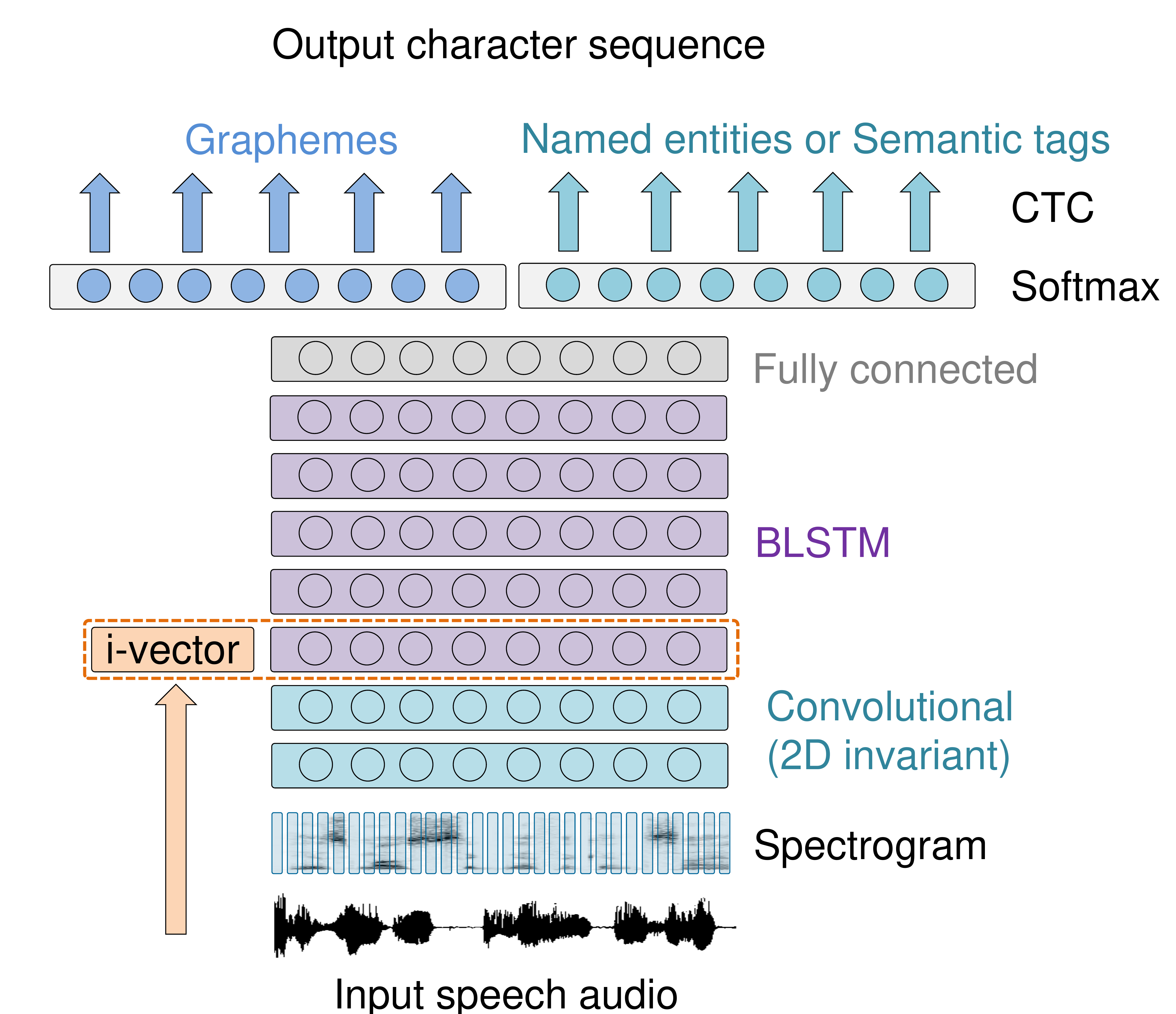}
\caption{Universal end-to-end deep neural network model  architecture for ASR, NER and SF tasks. 
Depending on the  task, the set of the output characters consists of: (1)~ASR: graphemes for a given language;  (2) NER:  graphemes and named entity tags;  (3) SF: graphemes and semantic SF tags.  }
\label{fig:ds}
\end{figure}

For   SLU   tasks, there is also an emerging interest in the end-to-end models which have a speech signal as input. 
Thus, acoustic, and particularly speaker, adaptation for such models can play an important role in improving the overall performance of these systems.  However, to our knowledge, there is no research on speaker adaptation for end-to-end SLU models, and the existing works do not use any speaker adaptation. 
One way to improve SLU models which we investigate in this paper is speaker adaptation.
%
 We apply i-vector based speaker adaptation~\cite{saon2013speaker}.
%
%
The proposed way of integration of i-vectors into the end-to-end model architecture is shown in Figure~\ref{fig:ds}. 
Speaker i-vectors are appended to the outputs of the last (second) convolutional  layer, just before the first recurrent (BLSTM) layer.
In this paper, 
for better initialization, we first train a model with \textit{zero pseudo i-vectors} (all values are equal to~0). Then, we use this pretrained model and fine-tune it on the same data but with the real i-vectors.
This approach was inspired by~\cite{deena2017semi}, where an idea of using zero auxiliary features during pretraining was implemented for language models and  in our preliminary experiments it demonstrated better results than direct model training with i-vectors~\cite{tomashenko2019investigating}.

\subsection{Transfer learning}\label{sec:trans_lern}
 Transfer learning  is a popular and  efficient method to improve the learning performance of  
 the target predictive function using  knowledge from a different source domain~\cite{pan2010survey}.
It allows to train  a model for a given target task using available out-of-domain  source data, and hence
to avoid an expensive data labeling process, which is especially useful in case of low-resource scenarios.

In this paper, for SF,  we investigate the effectiveness of  transfer learning   for various source domains and tasks: (1)  ASR in the target and  out-of-domain languages; (2)  NER in the target language; (3) SF.
%
%
%
%
For all the tasks, we used similar model architectures (Section~\ref{sec:models} and Figure~\ref{fig:ds}). 
The difference is in the text data preparation and output targets.
For training ASR systems, the output targets correspond to alphabetic characters and a \textit{blank}  symbol. 
For NER tasks, the output targets include all the ASR targets and targets corresponding to named entity tags. 
We have several symbols corresponding to named entities (in the text these characters are situated before the beginning of a named entity, which can be a single word or a sequence of several words) and  a one tag corresponding to the end of the named entity, which is the same for all named entities.
Similarly, for SF tags, we use targets corresponding to the semantic concept tags and one tag corresponding to the end of a concept.
Transfer learning is realized through the chain of consequence model training on different tasks.
For example, we can start from training an ASR model on audio data and corresponding text transcriptions. Then, we change the softmax layer in this model by replacing the targets with the SF targets and continue  training on the corpus annotated with semantic tags.  Further in the paper, we denote this type of chain as $ASR {\textcolor{blue}{\rightarrow}} SF$.
Models in this chain can be trained on different corpora, that can make this method   useful in low-resource scenario when we do not have enough semantically annotated data  to train an end-to-end model, but have sufficient amount of data annotated with more general concepts or only transcribed data.
For NER, we also investigates the knowledge transfer from ASR.


\begin{table}[tbh]
  \caption{Corpus statistics  for ASR, NER and SF tasks. }
  \label{tab:data}
  \label{tab:example}
  \centering
   \footnotesize
  \begin{tabular}{|l|l|l|l|}
\hline
   \textbf{Task} & \textbf{Corpora} &  \textbf{Size,h} & \textbf{\#Speakers}  \\
    \hline
  {ASR train} & \scriptsize EPAC, ESTER~1,2, ETAPE, REPERE,  DECODA, MEDIA, PORTMEDIA & 404.6  & 12518    \\
          \hline \hline 
  {NER train} & \scriptsize EPAC, ESTER 1,2, ETAPE, REPERE & 323.8  & 7327\\
   \hline
{NER dev } &\scriptsize  ETAPE \footnotesize (dev) & 6.6  & 152    \\ 
 \hline
  NER test& \scriptsize ETAPE  \footnotesize(test),  Quaero \footnotesize(test) & 12.3  & 474    \\ 
   \hline \hline 
  SF train                 & 1. \scriptsize MEDIA  \footnotesize (train), & 16.1   & 727      \\ \cline{2-4}
                          & 2. \scriptsize PORTMEDIA  \footnotesize (train)  & 7.2  &  257       \\  \cline{1-4}
  SF   dev               &\scriptsize  MEDIA  \footnotesize (dev) & 1.7   & 79     \\ \cline{1-4}
  SF   test              & \scriptsize MEDIA  \footnotesize (test) & 4.8   & 208  
    \\
    \hline
  \end{tabular}
\end{table}


\section{Experiments}\label{sec:exp}


\subsection{Data}
\label{ssec:data}


Several publicly available corpora have been used for experiments (see Table~\ref{tab:data}).


\subsubsection{ASR data}


The corpus for ASR training was composed of  corpora from
various evaluation campaigns in the field of automatic speech processing for
French. 
The EPAC~\cite{esteve2010epac}, ESTER~1,2~\cite{galliano2009ester}, ETAPE~\cite{gravier2012etape},  REPERE~\cite{giraudel2012repere} contain transcribed speech in French from TV and  radio broadcasts.
These data were originally in the microphone channel and for experiments in this paper were  downsampled from 16kHz to 8kHz, since the test set for our main target task (SF) consists of  telephone conversations. The DECODA~\cite{bechet2012decoda} corpus is composed of  dialogues from the  call-center 
of the Paris transport authority. The  MEDIA~\cite{devillers2004french,bonneau2006results} and PORTMEDIA~\cite{lefevre2012robustesse}  are corpora of dialogues
simulating a vocal tourist information server.
The target language in all experiments is French.
For experiments with  transfer learning  from ASR built in a different source language  to SF in the target language,
we used the TED-LIUM corpus~\cite{rousseau2014enhancing}. This
publicly available dataset contains 1495 TED talks in English that amount
to 207 hours of speech
data from 1242 speakers.

\vspace{-2pt}

\subsubsection{NER data}

To train the NER system, we used the following corpora: 
EPAC, ESTER~1,2,  ETAPE, and REPERE. 
These corpora contain speech with text transcriptions and named entity annotation. The named entity annotation is performed following the methodology of the  Quaero project~\cite{grouin2011proposal}.
The taxonomy is composed of 8 main types: \textit{person, function, organization, location, product,
amount, time}, and \textit{event}. Each named entity can be a single word or a sequence of several words. 
The total amount of annotated data is 112 hours. Based on this data, a classical NER system was trained using \textit{NeuroNLP2}\footnote{https://github.com/XuezheMax/NeuroNLP2} to automatically extract named entities for the rest 212 hours of the training corpus. This was done in order to increase the amount of the training data for NER.
Thus, the total amount of audio  data to train the NER system is about 324  (112+212) hours.
The development part of the ETAPE corpus was used for development, and as a test set we used the ETAPE test and Quaero test datasets. 

\subsubsection{SF data}

The  following two  French corpora, dedicated to semantic extraction from speech in a context of human/machine dialogues, were used in the current experiments: MEDIA and PORTMEDIA.
The corpora have  manual transcription and conceptual annotation~\cite{vukotic2015time,devillers2004french}.
The MEDIA corpus is related to the hotel booking domain, and its annotation contains $76$ semantic tags: \textit{ room number, hotel name, location, date}, etc.
The PORTMEDIA corpus is related to the theater ticket reservation domain and its annotation contains $35$ semantic tags which are very similar to the tags used in the MEDIA corpus. 
For joint training on these corpora, we used a combined set of $86$ semantic tags.


\subsection{Models}\label{sec:models}

We used the \textit{deepspeech.torch} implementation\footnote{https://github.com/SeanNaren/deepspeech.pytorch}  for training speaker independent (SI) models, and our modification of this implementation to integrate speaker adaptation.
The open-source \textit{Kaldi} toolkit~\cite{povey2011kaldi} was used to extract  100-dimensional speaker  i-vectors.
All models had similar topology (except for the number of outputs) shown in Figure~\ref{fig:ds} for SAT models. SI models were trained in the same way, but without i-vector integration. 
Input features are spectrograms. 
They are followed by two 2D-invariant (in the time and-frequency domain) convolutional layers\footnote{With parameters: kernel size=(41, 11), stride=(2, 2), padding=(20, 5)}, and then by five 800-dimensional BLSTM layers with sequence-wise batch normalization.   A fully connected layer is applied after BLSTM layers,  and the output layer of the neural network is a softmax layer.  The size of the output layer depends on the task (see Section~\ref{sec:tasks}).
The model is trained using the CTC loss function.

\subsection{Tasks}\label{sec:tasks}

The target tasks for us are NER and SF.
For each of this task, other tasks can be used for knowledge transfer.
To train NER, we use ASR for transfer learning.
To train SF, we use ASR on French and English, NER and another auxiliary SF task for transfer learning.
Hence, we consider the following set of tasks:
%
\begin{itemize}
\footnotesize
    \item{$ASR_F$} -- French ASR with 43 outputs \{French characters, \textit{blank} symbol\}.
    \item{$ASR_E$} --   English ASR with 28 outputs \{English  characters,  \textit{blank} symbol\}.
    \item{$NER$} -- French NER with 52  outputs \{43 outputs from $ASR_F$, 8 outputs corresponding to named entity tags, 1 output corresponding to the closing tag for all named entities\}.
    \item{$SF_1$} -- target SF task with 130 outputs \{43 outputs from $ASR_F$ , 86 outputs  for semantic slot tags, 1 output for the closing tag\}; trained on the training part of the MEDIA corpus. 
    \item{$SF_{1+2}$} -- auxiliary SF task; trained on the MEDIA plus  PORTMEDIA training corpora.  
    \item{$NER^\star$, $SF_1^\star$} -- for the target tasks $NER$ and $SF_1$, we also considered $\star$-mode (Section~\ref{sec:starmode}).
\end{itemize}
\vspace{-3pt}

\subsection{Results for NER}\label{sec:res_ner}

Performance of NER was evaluated in terms of \textit{precision}, \textit{recall}, and \textit{F-measure}. 
Results for different training chains for speaker-independent (SI) and speaker adaptive training models (SAT) are given in Table~\ref{tab:ner_res}. We can see, that pretraining with $ASR_F$ task does not lead to significant improvement in performance.
When the $NER^\star$ is added to the training chain, it improves all the evaluation measures. In particular, F-measure is increased by 1.9\% absolute.
For each  training chain, we trained a  corresponding chain with speaker adaptation. Results for SAT models are given in the right part of Table~\ref{tab:ner_res}. For all training chains, SAT models outperform SI models. The best result with SAT (F-measure 71.8\%) outperforms the best SI result by 1.1\% absolute.

\begin{table}
\centering
\caption{NER results on the test dataset in terms of Precision (P,\%), Recall (R,\%) and F-measure (F, \%) for SI and SAT models.}\label{tab:ner_res}
\begin{tabular}{|l|l|l|l|||l|l|l|}
\hline
\multicolumn{1}{|c|}{\textbf{Model training}} &   \multicolumn{3}{c|||}{\textbf{SI}} & \multicolumn{3}{c|}{\textbf{SAT}} \\ \cline{2-7}
 &   \textbf{P} & \textbf{R} & \textbf{F} & \textbf{P} & \textbf{R} & \textbf{F}\\ \hline    
$NER$   & 78.9 & 60.7 &68.6     & 80.9 & 60.9 &69.5 \\
$ASR_F{\textcolor{blue}{\rightarrow}}NER$ &  80.5 & 60.0 & 68.8 & 80.2 & 61.7 & 69.7 \\
$ASR_F{\textcolor{blue}{\rightarrow}}NER{\textcolor{blue}{\rightarrow}}NER^\star$  & 82.1 & 62.1 & 70.7 & \textbf{83.1} & \textbf{63.2} & \textbf{71.8}\\  \hline
\end{tabular}
\end{table}

\vspace{-18pt}
\subsection{Results for SF}\label{sec:res_sf}
\begin{figure}[tph]
\centering\includegraphics[width=87mm]{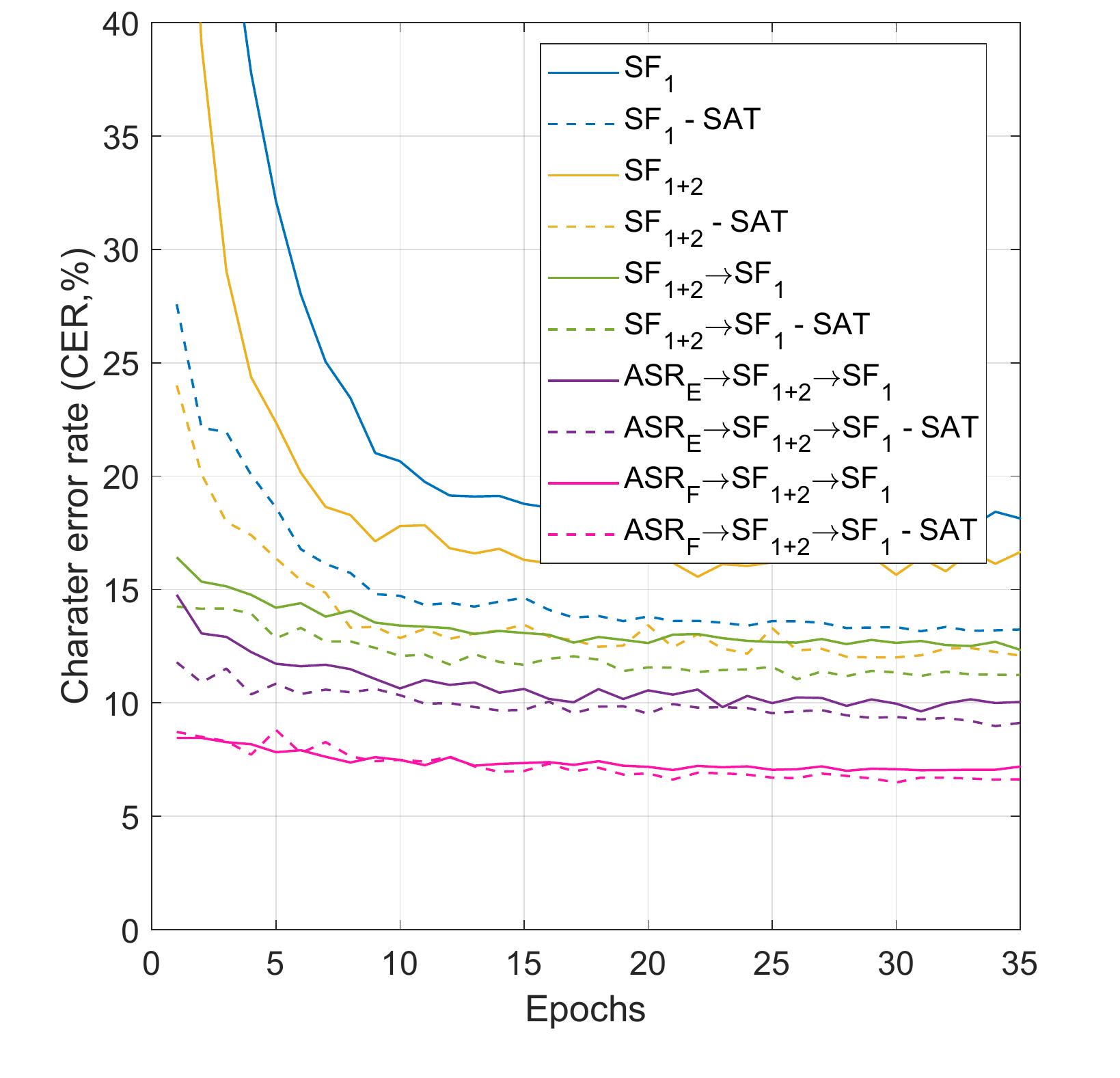}
\caption{Training performance on the MEDIA development dataset in terms of character error rate (CER) for training SF models. For each type of the model chain, a solid line corresponds to a SI model, and a dash line of the same colour denotes a SAT version of a given model.}
\label{fig:curves}
\end{figure}

SF performance  was evaluated in terms of \textit{F-measure}, \textit{concept error rate} (CER) and \textit{concept value error rate} (CVER).
Training performance on the MEDIA development dataset in terms of \textit{character error rate} (CER) is shown in Figure~\ref{fig:curves} for different transfer learning chains for  SI and SAT models.
The blue curves $SF_1$  corresponds to the SI baseline model when the model was directly trained on the target SF task without pretraining. 
All  curves of other colours correspond to different sequential transfer learning chains. 
All considered transfer learning schemes substantially improve the training performance. 
By comparing $SF_1$ and $SF_{1+2}$, we can conclude that training on the auxiliary task improves the performance. When we further trained this model on the target task ($SF_{1+2}{\rightarrow}SF_1$), the performance continued to improve. This demonstrates, that in given conditions,  the sequence transfer learning provides better improvement than just joint training.
The best SI model is obtained through the following training chain:  $ASR_F{\rightarrow}SF_{1+2}{\rightarrow}SF_{1}$.
These results are confirmed further in Table~\ref{tab:res_sf}.
Also, we can see that SAT gives an additional improvement in performance  for all the models.

\begin{table}[hpt]
 \caption{ {SF performance results on the MEDIA test dataset for  end-to-end SF models trained with different transfer learning approaches. Results are given  in terms of  F-measure (F),  CER and CVER metrics (\%);  \textbf{SF$_1$} -- target task;   \textbf{SF$_{1+2}$}  -- auxiliary task; \textbf{F} and \textbf{E} refer to the languages.  For the best models,  the results in blue  correspond to  decoding using beam search with a LM.}\label{tab:res_sf}}
  \centering
   \begin{tabular}{|l|||l|l|l|l|||l|l|l|l| }
   \hline
\multicolumn{1}{|c|||}{\textbf{Model training}}  & \multicolumn{4}{c|||}{\textbf{SI}} & \multicolumn{4}{c|}{\textbf{SAT}}\\ \cline{2-9}
 & \textbf{\#~~}  & \textbf{F~~~~~} & \textbf{CER} & \textbf{CVER} & \textbf{\#} & \textbf{F \textcolor{blue}{(LM)}} & \textbf{CER \textcolor{blue}{(LM)}} & \textbf{CVER \textcolor{blue}{(LM)}}\\ 
   \hline
 $SF_1$                                      &  1 & 72.5	&	39.4 & 52.7&\multicolumn{4}{c|}{}\\
$SF_{1+2}$                                    & 2     &  73.2	&	39.0 &50.1 &\multicolumn{4}{c|}{}\\ 
 $SF_{1+2}{\textcolor{blue}{\rightarrow}}SF_1$  & 3   &  77.4	&	33.9 & 44.9&\multicolumn{4}{c|}{}\\
 $ASR_{E}{\textcolor{blue}{\rightarrow}}SF_{1+2}{\textcolor{blue}{\rightarrow}}SF_1$ & 4 &   81.3	&	28.4&37.3&\multicolumn{4}{c|}{}\\ \hline \hline
 $ASR_{F}{\textcolor{blue}{\rightarrow}}SF_{1+2}{\textcolor{blue}{\rightarrow}}SF_1$ & 5 &  85.9 	&	21.7  & 28.4& 9 & 87.5	&	19.4 & 25.4 \\ 
$NER{\textcolor{blue}{\rightarrow}}SF_{1+2}{\textcolor{blue}{\rightarrow}}SF_1$ 	        & 6 & 86.4	&	20.9 & 27.5& 10& 87.3 &	19.5 & 26.0  \\ \hline
 $ASR_{F}{\textcolor{blue}{\rightarrow}}SF_{1+2}{\textcolor{blue}{\rightarrow}}SF^\star_1$    &7 &  85.9 	&	21.2  & 27.9 & 11 &\textbf{87.7 \textcolor{blue}{(89.2)}}	&	18.8 \textcolor{blue}{(16.5)} & 25.5 \textcolor{blue}{\textbf{(20.8)}}\\
 $NER{\textcolor{blue}{\rightarrow}}SF_{1+2}{\textcolor{blue}{\rightarrow}}SF^\star_1$	         & 8 &87.1	&	19.5 & 27.0 & 12 & 87.6 \textcolor{blue}{(\textbf{89.2})} &	\textbf{18.6} \textcolor{blue}{(\textbf{16.2})} & \textbf{24.6} \textcolor{blue}{(\textbf{20.8})}\\ \hline 
  \end{tabular}
\end{table}

\begin{table}[hpt]
\footnotesize
 \caption{ {SF performance results on the MEDIA test dataset for different systems.}\label{tab:res_resume}}
  \centering
   \begin{tabular}{|lr|||lr| }
   \hline
\textbf{Systems in literature:}  & \textbf{~~CER~~} & \textbf{Systems  in this paper:} & \textbf{~~CER~~}\\ \hline
Pipeline: ASR+SLU,~\cite{simonnet2018simulating}  & 19.9~~ &  ---gready mode & 18.6~~ \\
End-to-end,~\cite{ghannay2018end} & 27.0~~ & ---beam search with LM & 16.2~~ \\\hline
  \end{tabular}
\end{table}

Results for different training chains for  speaker-independent (SI) models on the test set are given in Table~\ref{tab:res_sf} (\#1--8). 
The first line $SF_1$ shows the baseline result on the test MEDIA dataset for the SF task, when a model was trained directly on the target task using  in-domain data for this task  (training part of the MEDIA corpus). 
The second line $SF_{1+2}$ corresponds to the case when the model was trained on the auxiliary SF task.
Other lines in the table correspond to different training chains described in Section~\ref{sec:trans_lern}. 
In  $\#4$, we can see a chain  
that starts from training an ASR model for English. We can observe that using  a pretrained ASR model from a different language can 
significantly 
($16.2\%$ of relative CER reduction) improve the performance of the SF model (\#4 vs \#3).
This result is noticeable since it shows that we can take benefit from linguistic resources from another language in case of lack of data for the target one.
Using an ASR model trained in French 
(\#5) provides  better improvement:   $36.0\%$ of relative CER reduction  (\#5 vs \#3).
When we start the training process from a NER model (\#6) we can observe slightly better results.
Further, for the best two model training chains (\#5 and 6) we trained corresponding models in $\star$-mode (\#7 and 8).
Results with speaker adaptation for four best models are shown in the right part  of Table~\ref{tab:res_sf} (\#9--12). We can see that SAT models show better results than  SI ones. 
For CVER,  we can observe a similar tendency. 
The results for the best models using beam search and a 4-gram LM are shown in brackets in blue. The LM was
built on the texts including "$\star$".
Finally, Table~\ref{tab:res_resume} resumes our best results (in greedy and beam search modes) and shows the comparison results on the MEDIA dataset from other works~\cite{simonnet2018simulating,ghannay2018end}.
We can see, that the reported results significantly outperform the results reported in the literature for the current task.


\begin{figure}[tph]
\centering\includegraphics[width=80mm]{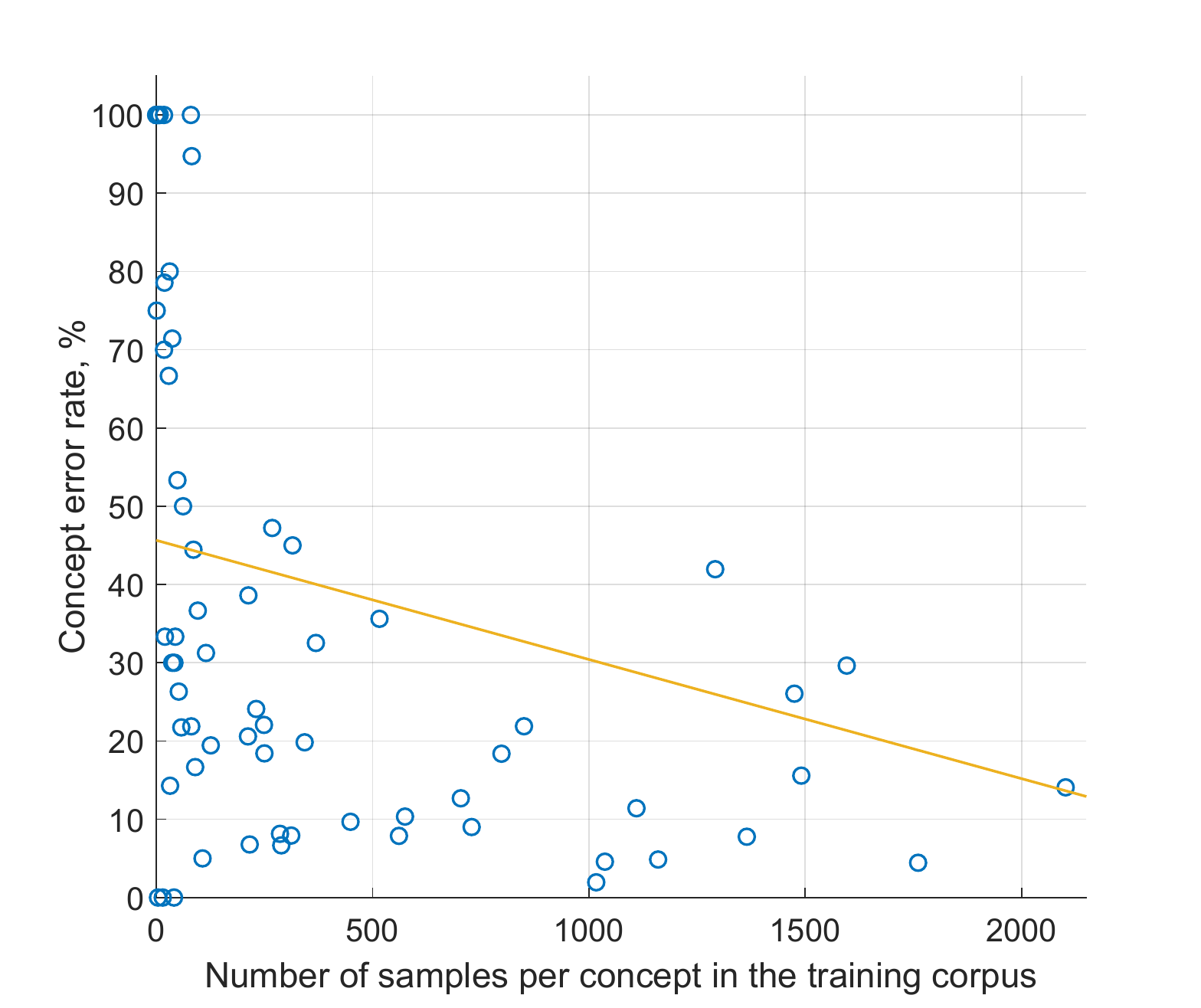}
\caption{Concept error rate (CER,\%) results on the MEDIA test dataset for different concepts depending on the number of corresponding concepts in the training corpus. The CER results are given for the SAT model (\#12), decoding with beam search and a 4-gram LM.}
\label{fig:cer_vs_freq}
\end{figure}

\subsubsection{Error analysis}

In the training corpus, different semantic concepts have different number of samples, that may impact the SF performance. Figure~\ref{fig:cer_vs_freq} demonstrates the relation between the concept error rate (CER) of a particular semantic concept  and its frequency in the training corpus. Each point in  Figure~\ref{fig:cer_vs_freq} corresponds to a particular semantic concept. For  rare tags, the distribution of errors has  larger variance and means than for more frequent tags.
\begin{figure}[tph]
\centering\includegraphics[width=122mm]{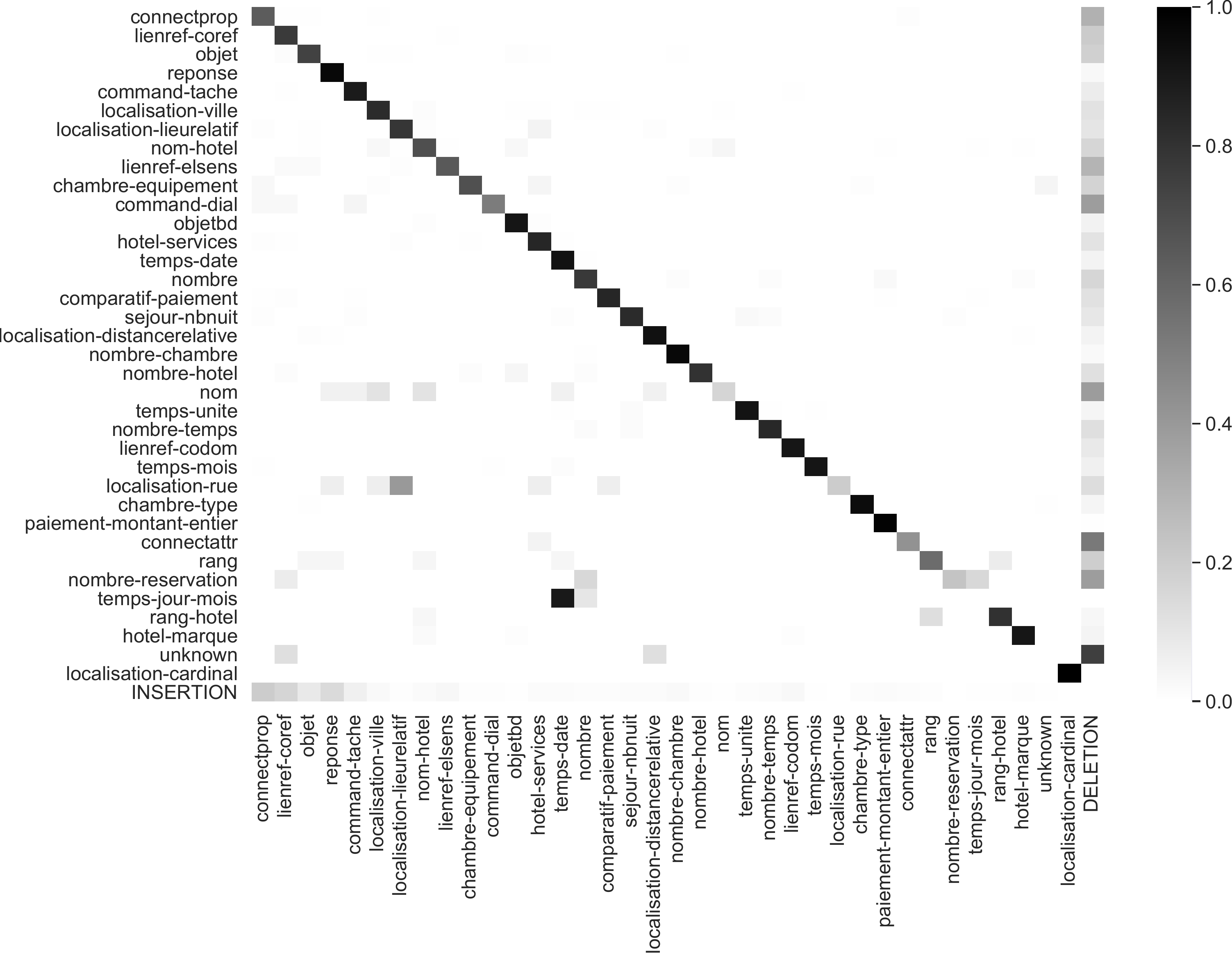}
\caption{Confusion matrix for concepts on the MEDIA test dataset. The last row and last column represent insertion and deletion errors correspondingly.  The CER results are given for the SAT model (\#12), decoding with beam search and a 4-gram LM.}
\label{fig:hp}
\end{figure}

In addition, we are interested in the distribution of different types of SF errors (\textit{deletions}, \textit{insertions} and \textit{substitutions}), which is shown in the form of a confusion matrix in Figure~\ref{fig:hp}. For better representation, we first ordered the concepts in descending order  by the total number of errors. Then, we chose the first 36 concepts which have the biggest number of errors. The total amount of errors of the chosen 36 concepts corresponds to 90\% of all the errors for all concepts in the test MEDIA dataset.
The diagonal corresponds to the  correctly detected concepts and  other elements (except for the last row and last column) correspond to  the substitution errors. The final raw represents insertion errors and the final column -- deletions.
Each element in the matrix shows  the total number of the corresponding events (\textit{correctly recognized concept, substitution, deletion or insertion})  normalized by the total number of such events in the row. 
The most frequent errors are deletions (50\% of all errors), then substitutions (32.3\%) and insertions (17.7\%).


\section{Conclusions}\label{sec:concl}

\vspace{-5pt}

In this paper, we have investigated 
several ways to improve the performance of end-to-end SLU systems.
We demonstrated
the effectiveness of speaker adaptive training and various   transfer  learning approaches for two end-to-end SLU tasks: NER and SF.
In order to improve the quality of the SF models, during the training, we proposed to use 
knowledge transfer  from an ASR system in another language and  from a NER in a target language. 
Experiments on the French MEDIA test corpus demonstrated that using  knowledge transfer from the ASR in English  improves the SF model performance by about 16\% of relative CER reduction for SI models.  

The improvement from the transfer learning is greater when the ASR model is trained on the target language (36\% of relative CER reduction) or when the NER model in the target language is used for pretraining.
Another contribution concerns SAT training for SLU models --
we  demonstrated that this can significantly improve the model performance for NER and SF.


\section{Acknowledgements}
This work was supported by the French ANR Agency through the ON-TRAC project, under the contract number ANR-18-CE23-0021-01, and by the RFI Atlanstic2020 RAPACE project.

\bibliographystyle{splncs04}
\bibliography{main}

\end{document}